\documentclass[10pt,times,mathptm,conference]{ieeeconf2}
\usepackage{amsmath,graphicx}
\IEEEoverridecommandlockouts  

\title{\LARGE \bf
Segmentation of both Diseased and Healthy Skin \\ from Clinical Photographs in a Primary Care Setting
}

 \author{ \em  Noel C. F. Codella$^{1}$,  Daren Anderson$^{2}$,   Tyler Philips$^{2}$, Anthony Porto$^{2}$, Kevin Massey$^{2}$, \\ Jane Snowdon$^{3}$, Rogerio Feris$^{1}$,  John Smith$^{1}$
 \thanks{$^{1}$ IBM T.J. Watson Research Center, Yorktown Heights, NY, USA. {\tt\small \{nccodell, rferis, jsmith\}@us.ibm.com}}  
 \thanks{$^{2}$ Community Health Center Inc., Middletown, CT, USA. {\tt\small \{andersd, phillipt, portoa, masseyk\}@chc1.com} } 
 \thanks{$^{3}$ IBM Watson Health, Yorktown Heights, NY, USA. {\tt\small \{snowdonj\}@us.ibm.com}}
\thanks{Accepted for publication at EMBC 2018. \copyright 2018 IEEE}
 }

\begin{document}

%
\maketitle
%
\thispagestyle{empty}
\pagestyle{empty}

\begin{abstract}
This work presents the first segmentation study of both diseased and healthy skin in standard camera photographs from a clinical environment. Challenges arise from varied lighting conditions, skin types, backgrounds, and pathological states. For study, 400 clinical photographs (with skin segmentation masks) representing various pathological states of skin are retrospectively collected from a primary care network. 100 images are used for training and fine-tuning, and 300 are used for evaluation. This distribution between training and test partitions is chosen to reflect the difficulty in amassing large quantities of labeled data in this domain. A deep learning approach is used, and 3 public segmentation datasets of healthy skin are collected to study the potential benefits of pre-training. Two variants of U-Net are evaluated: U-Net and Dense Residual U-Net. We find that Dense Residual U-Nets have a 7.8\% improvement in Jaccard, compared to classical U-Net architectures (0.55 vs. 0.51 Jaccard), for direct transfer, where fine-tuning data is not utilized. However, U-Net outperforms Dense Residual U-Net for both direct training (0.83 vs. 0.80) and fine-tuning (0.89 vs. 0.88). The stark performance improvement with fine-tuning compared to direct transfer and direct training emphasizes both the need for adequate representative data of diseased skin, and the utility of other publicly available data sources for this task.
\end{abstract}
%

\section{Introduction}
\label{sec:intro}

Segmentation of skin may be an important pre-processing step for any system that is designed to interact with the patient or extract physiological insights from images of the patient. Disease classification decisions can be influenced by the quality of segmentation. In dermoscopic images, segmentation masks have demonstrated significant influence in the performance of disease classification systems \cite{codellajrd, yuisbi}. Recent challenges hosted at the International Symposium of Biomedical Imaging (ISBI) in 2016 and 2017 around the analysis of dermoscopic images have devoted an entire task toward segmentation \cite{challenge}.

\begin{figure}[t]
  \centerline{\includegraphics[width=8cm]{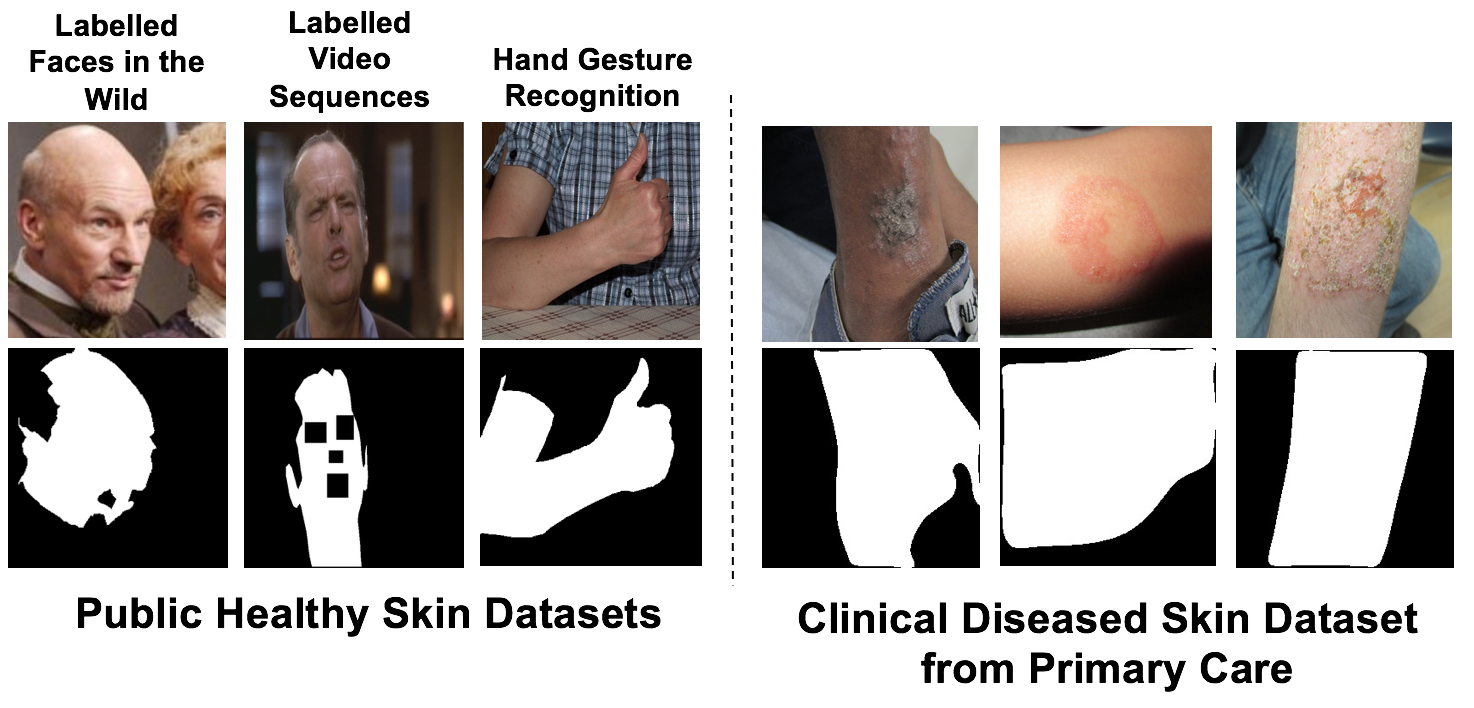}}
  \caption{Datasets used in this work. {\em Top:} Original images. {\em Bottom:} Corresponding ground-truth segmentation masks. Existing public datasets of skin segmentation (left) are not representative of both healthy and diseased skin seen in primary care clinical photography (right).  }
\label{fig:data}
\end{figure}

While segmentation of healthy skin from standard camera images has also been the subject of prior research \cite{datasetlfw,datasetbumovie,datasethandgesture1,datasethandgesture2, datasethandgesture3}, little work to date has focused on skin segmentation in situations where skin may be diseased, such as standard camera images used for documentation and diagnosis in a primary care setting (Fig. ~\ref{fig:data}). Most recent papers on classification of dermatological conditions from standard cameras have ignored the task of skin segmentation, relying at least in part on the data being pre-localized by the photographer \cite{nature,eccv}. 

\begin{figure*}[t]
  \centerline{\includegraphics[width=15cm]{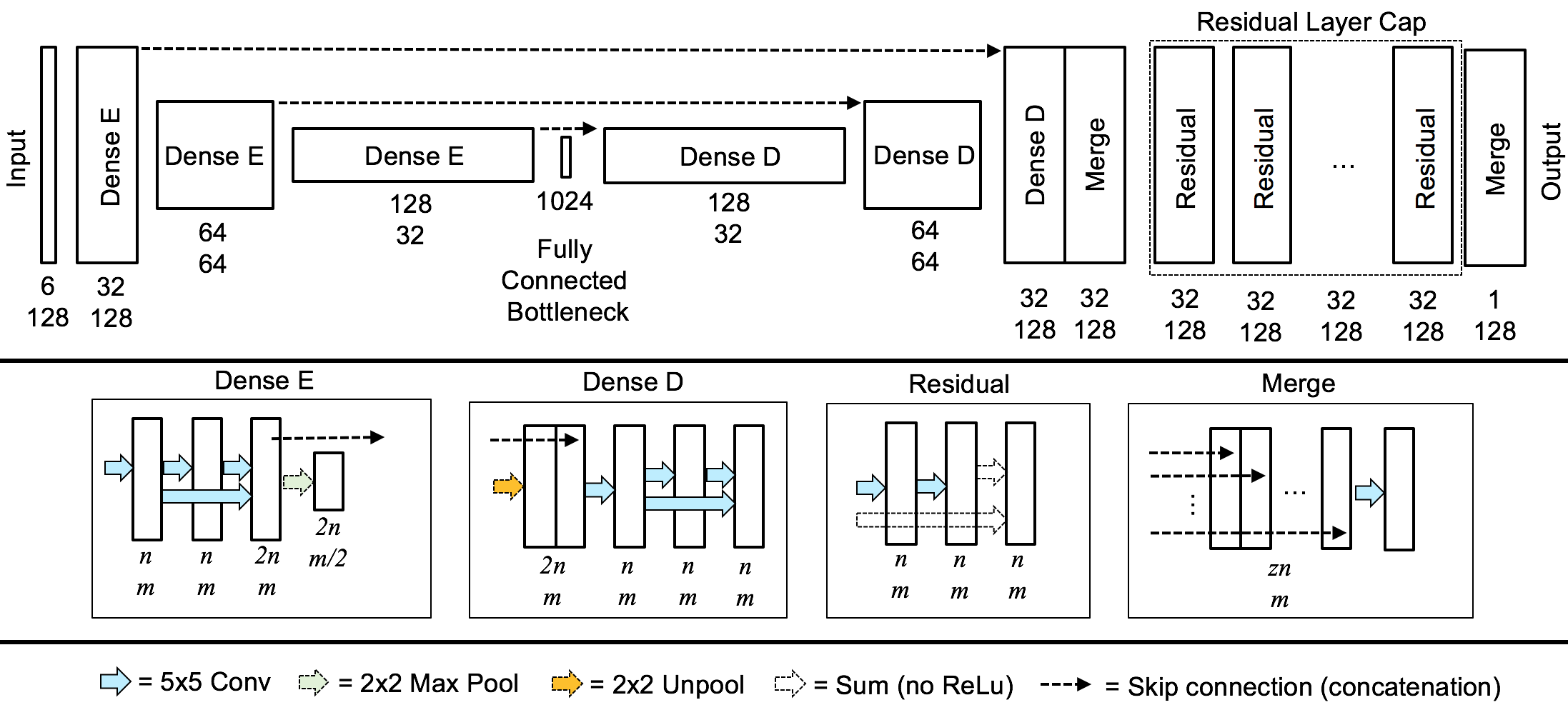}}
  \caption{Dense Residual U-Net Architecture. {\em Top:} Overview diagram of network structure. System begins with the U-Net architecture above the input layer (Red, Green, Blue, Hue, Saturation, and Value channels), where convolution blocks have been replaced with densely connected convolution blocks during encoding (E) and decoding (D) stages. This is followed by a merge layer, which concatenates all the individual convolution outputs from the last densely connected block, and then inputs these into a convolution layer. This is followed by consecutive residual blocks, and finally another merge layer, which takes as input the outputs of each of the prior residual blocks. Number of filters and spatial dimensionality listed below structural blocks.   {\em Center:} Details of Dense Encoding (E), Dense Decoding (D), Residual, and Merge network blocks. The number of convolution filters is denoted by {\em n}, and the spatial dimension is denoted by {\em m}. {\em z} denotes number of independent input layers into a merge layer. {\em Bottom:} Key denoting operations and their visual representations.  }
\label{fig:dru}
\end{figure*}



Segmentation, both within and outside the field of medical imaging, is a large and active field \cite{segreview1,segreview2,segreview3}, too broad to comprehensively capture here. However, most recent approaches involve deep learning, and have generally been applied with great success \cite{maskrcnn, fcn, unet, deepseg1, deepseg2}.  Fully convolutional networks \cite{fcn} applied to standard camera photos led to the development of the U-Net \cite{unet}, which attained top performance on the ISBI 2012 neuronal structure segmentation challenge. Since that time, similar structures have achieved top performance on a multitude of other segmentation tasks, including dermoscopic skin image segmentation \cite{codellajrd, yuisbi, challenge, lossfunction}.  

In this work, deep learning approaches are used for the first study of segmentation of both healthy and diseased skin in standard camera clinical photographs. While, as described, many datasets exist for healthy skin segmentation, none exist for diseased skin. The well-studied U-Net architecture is used as a baseline network design. Recent advances from research in classification, namely Densely Connected Networks \cite{densenet} and Deep Residual Networks \cite{resnet}, are used to study another U-Net variant: a Dense Residual U-Net, where convolutions are replaced with dense convolutions to increase feature re-use, and residual layers are added to the end of the U-Net to model contextual awareness of situations where the base U-Net may be making mistakes. Public image segmentation datasets for healthy skin are used to pre-train networks. A specialized dataset of 400 images is retrospectively collected from a primary care network and annotated, where 300 images are used for the purposes of evaluation, and 100 images are used to train and fine-tune. When fine-tuning data is not available, we find that Dense Residual U-Nets have a 7.8\% improvement in Jaccard, compared to classical U-Net architectures, suggesting an improvement in ability to generalize across domains.

\section{Network Architectures}

\subsection{Dense Residual U-Net}
The Dense Residual U-Net architecture is visually described in Fig. \ref{fig:dru}. It is similar to earlier descriptions of the U-Net \cite{unet}, with the following three changes: 1) convolutions are replaced with dense convolutions \cite{densenet} to increase feature re-use, 2) a series of 4 residual convolutions are appended to the output of the network to better model context and predict under what circumstances the base U-Net makes mistakes \cite{resnet}, 3) a fully-connected layer is added to the bottleneck to model global non-linear dependencies, which previous reports have found improved performance \cite{codellajrd, recod}. Rectified linear units are used as the nonlinearity for all neurons, except the final layer, which uses a hyperbolic tangent. Inputs are standard normalized per-image and per-channel. 

Image dimensions were resized to 128x128, batch size was set to 16, momentum 0.9, and training ran up to 500 epochs (early stopping occurs with 50 epochs of no improvement in 20\% validation). 50\% dropout was applied to the last dense encoding block, the fully connected layer, and all dense decoding blocks. Gaussian noise with 0 mean and 0.025 standard deviation was added prior to each dense encoding block. Standard data augmentation of rotation, scale, and cropping was applied. For training from random network weight initialization, a learning rate of 0.01 was used. For fine-tuning, this was reduced to 0.001. The objective function implemented was designed to correlate with the Jaccard index, and has been described in previous work \cite{lossfunction}:
\begin{equation}
L_{i,j} = 1 - \frac{\sum_{i,j}(t_{i,j} p_{i,j})}{\sum_{i,j}(t_{i,j}^{2})+\sum_{i,j}(p_{i,j}^{2}) - \sum_{i,j}(t_{i,j} p_{i,j})}
\end{equation} 
where $L{(i,j)}$ represents the loss at pixel location {\em (i,j)} in an image, {\em $t_{i,j}$ } is the ground truth mask value (skin 1, non-skin 0) for that pixel, and {\em $p_{i,j}$ } is a value between 0 and 1 representing the confidence of the network that the pixel is skin (re-scaled hyperbolic tangent). For binary decisions, a threshold of 0.5 (50\% confidence) is applied to all pixels. 

\subsection{Traditional U-Net}

A U-Net is a similar network structure as Dense Residual U-Net, with the residual end-units and dense connections removed \cite{codellajrd,unet}. Two variants are studied: one with the same number of filters (termed simply ``U-Net''), and one where the dimensionality of the fully connected layer and convolution filters are increased (to 1450 for fully connected, and 45, 90, and 180 filters for convolutions), to better match the number of total network parameters of the Dense Residual U-Net (termed ``U-Net Large''). 

\subsection{Training Scenarios}

{\em Direct Transfer} applies the pretrained network directly to evaluation task with no fine-tuning. {\em Fine-Tuning} uses the specified number of examples to fine-tune the pretrained network. {\em Direct Training} used a randomly initialized model to train from same data used to fine-tune.

\section{DATASETS}
\label{sec:datasets}


\subsection{Public Skin Datasets for Pre-Training}

\begin{figure*}[th!]
  \centerline{\includegraphics[width=15cm]{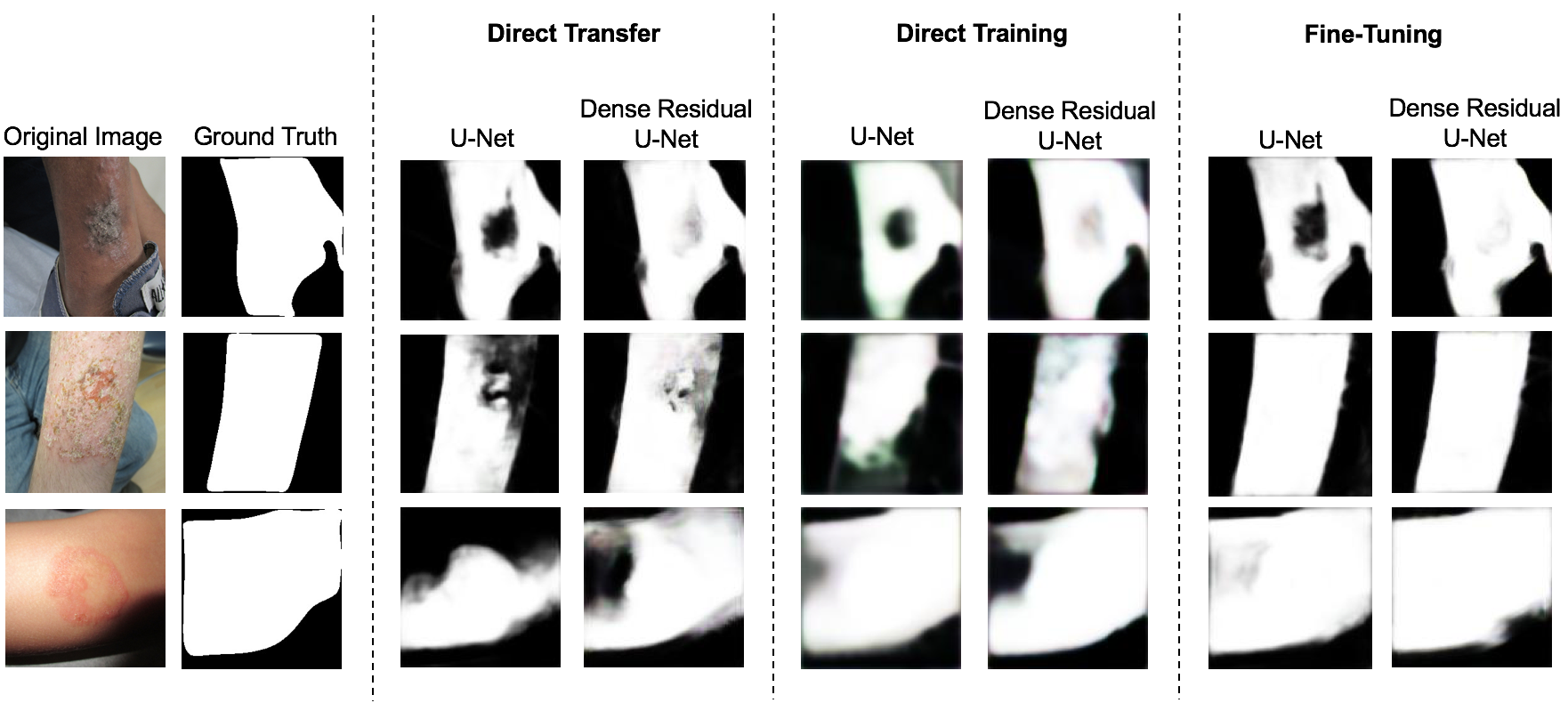}}
  \caption{Example network outputs on 3 clinical photographs. Raw confidence scores are shown (pixel intensity 0 is 0\%, and 255 is 100\%). Binary decision threshold is 50\% confidence.    }
\label{fig:results}
\end{figure*}

Public images with manually annotated skin regions were collected from 3 datasets (Fig. \ref{fig:data}.): 1) Labeled Faces in the Wild (LFW) \cite{datasetlfw}, 2) Labeled Video Sequences (LVS) \cite{datasetbumovie}, and 3) a hand gesture recognition (HGR) dataset \cite{datasethandgesture1, datasethandgesture2, datasethandgesture3}. Images from each dataset were randomly downsampled to roughly balance the sets (1000 from LFW, 721 from LVS, and 899 images from hand gesture, 2620 total).  

\subsection{Primary Care Clinical Photography Dataset}

Images were collected in collaboration with Community Health Center, Inc. (CHCI) a Federally Qualified Health Center (FQHC), caring for over 140,000 under-served patients across the state of Connecticut.  CHCI sought to improve access to specialty care while improving efficiency and reducing costs by implementing an eConsult system. The eConsult system allows primary care physicians (PCPs) to generate a patient report consisting of free-text case descriptions and photographs. This information can then be shared with a specialist, in order to help determine diagnosis, treatment plan, as well as whether an in-person visit with a specialist is necessary.

From the eConsult system, a subset of 400 retrospective clinical focused images were collected between March 2015 and March 2017 from a total set of 3405, according to MD5 sorted order. All data was de-identified according to HIPAA protocol, with Institutional Review Board (IRB) approval. The skin was manually traced with splines by a single user in each image to generate a binary mask. 300 of the images were used for evaluation, and 100 images were used for training and fine-tuning. This distribution is chosen to reflect the difficulty in amassing large quantities of labeled data in this domain.

As data was collected from a real-world primary care setting, the scope of captured disease classes was broad. Some of the most common physician reported classes of pathologies included various types of rashs (skin eruptions, dermatitis, dyshidrosis), neoplasms, acne, psoriasis, alopecia, and dyschromias. Other represented diseases, though less common, included warts, herpes, scabies, lichen, hemagiomas, nail disorders, and alopecia.

\section{RESULTS}
\label{sec:results}

\begin{table}[t!]
\centering
\begin{tabular}{|p{1.4cm}|p{1.2cm}|p{1.3cm}|p{1.3cm}|p{1.3cm}|p{1.3cm}|p{1.3cm}|p{1.2cm}|p{1.2cm}|} 
 \hline
 \bf Model Type & \bf \# Param / Epoch Time & \bf Direct Transfer & \bf Direct Training & \bf Fine-Tuning \\ 
 \hline
 {\bf Dense Residual U-Net} & 149.9M / 680 s & \bf 0.55 (0.71) & 0.80 (0.89)  & 0.88 (0.94)  \\ 
 \hline

 \bf U-Net Large & 150M / 697 s & 0.51 (0.67)  & \bf 0.83 (0.91) & \bf 0.89 (0.94) \\
 \hline
 \bf U-Net & 7.5M / 363 s & 0.44 (0.61) & 0.79 (0.88) & 0.89 (0.94)  \\
 \hline
\end{tabular}
\caption{Jaccard index (and Dice coefficient) for each network model and training scenario  }
\label{table:results}
\end{table}

Visual segmentation results are shown in Fig. \ref{fig:results}, and objective performance measures (according to Jaccard index and Dice coefficient) are listed in Table \ref{table:results}. 

Dense Residual U-Net demonstrated significant improvements over traditional U-Net when networks trained on public data sources were directly applied and evaluated on primary care clinical photos, even when accounting for the number of network parameters and training time. This suggests that the network structure may have improved capability to generalize across dataset domains and distributions. As seen in Fig. \ref{fig:results}, the Dense Residual U-Net demonstrated improved ability particularly to identify diseased areas of skin that deviated significantly from healthier regions. 

In the cases of direct training from random initialization of network parameters, this work confirms earlier reports \cite{unet} that these classes of network structures are robust to training on small dataset sizes: training occurred on 100 images, and yet reasonable results that outperformed direct transfer were obtained. 

In the cases of fine-tuning, all networks perform similarly well according to Jaccard index, and significant performance improvements were observed in comparison to direct training and direct transfer. While this result is to be expected, it first emphasizes the need to collect representative datasets in the domain of diseased skin, and second demonstrates that other skin segmentation datasets, while not representative of disease states, still provides useful information. 

According to subjective assessment, even though U-Net slightly outperformed Dense Residual U-Net in the fine-tuning experiment according to Jaccard, the segmentation results of Dense Residual U-Net appeared to be of higher quality. For example, the framework was able to better model some extreme lighting scenarios or diseased skin patches (Fig. ~\ref{fig:results}). This may be due to better generalization capabilities, also consistent with the direct transfer experiment, that improves subjective segmentation quality, but at a cost of slightly decreased overall performance. 

\section{CONCLUSION}
\label{sec:conclusion}

This work presents the first segmentation study of both healthy and diseased skin from standard camera clinical photographs in a primary care setting. While much work has been performed for segmentation of healthy skin, no attempt has been made to quantify how well methods trained on healthy skin translate to diseased skin, or how specific training or transfer learning may improve performance. An ability to accurately discern diseased skin from background may be critical to further downstream analysis, such as skin disease classification or extraction of other relevant physiological signals. 

Public data sources of healthy skin are used for pre-training. Two variants of a state-of-art deep learning approach for segmentation, the U-Net and Dense Residual U-Net, are evaluated.  A specialized dataset of 400 images was collected from a primary care network and annotated for the purposes of evaluation (300 images) as well as direct training and fine-tuning (100 images).

Relying on pre-training alone, the Dense Residual U-Net demonstrates an improvement of 7.8\% in Jaccard index over U-Net with a similar number of network parameters. In the cases of direct training from random initialization, this work confirms earlier reports \cite{unet} that demonstrate these classes of network structures are incredibly robust to training on small dataset sizes. In the cases of fine-tuning, all networks perform similarly well, and demonstrate significant improvements over both relying on pre-training alone, and direct training from random initialization. 

As dataset bias is always present in training data, and is difficult to quantify, the presented Dense Residual U-Net method holds potential to improve segmentation of skin in standard camera clinical photos where unexpected or rare diseases may present.


\bibliographystyle{IEEEbib}

%
%




\end{document}